# Gaze to Insight: A Scalable AI Approach for Detecting Gaze Behaviours in Face-to-Face Collaborative Learning

Junyuan Liang[1] [0000-0001-9119-2885], Qi Zhou[1] [0000-0002-4694-4598], Sahan Bulathwela[2] [0000-0002-5878-2143], and Mutlu Cukurova[1 2] [0000-0001-5843-4854]

[1] UCL Knowledge Lab, University College London, London, UK
[2] UCL Center for Artificial Intelligence, University College London, UK
{zczqjli, qtnvqz3,m.bulathwela,m.cukurova}@ucl.ac.uk

**Abstract.** Previous studies have illustrated the potential of analysing gaze behaviours in collaborative learning to provide educationally meaningful information for students to reflect on their learning. Over the past decades, machine learning approaches have been developed to automatically detect gaze behaviours from video data. Yet, since these approaches often require large amounts of labelled data for training, human annotation remains necessary. Additionally, researchers have questioned the cross-configuration robustness of machine learning models developed, as training datasets often fail to encompass the full range of situations encountered in educational contexts. To address these challenges, this study proposes a scalable artificial intelligence approach that leverages pretrained and foundation models to automatically detect gaze behaviours in face-to-face collaborative learning contexts without requiring human-annotated data. The approach utilises pretrained YOLO11 for person tracking, YOLOE-26 with text-prompt capability for education-related object detection, and the Gaze-LLE model for gaze target prediction. The results indicate that the proposed approach achieves an F1-score of 0.829 in detecting students' gaze behaviours from video data, with strong performance for laptop-directed gaze and peer-directed gaze, yet weaker performance for other gaze targets. Furthermore, when compared to other supervised machine learning approaches, the proposed method demonstrates superior and more stable performance in complex contexts, highlighting its better cross-configuration robustness. The implications of this approach for supporting students' collaborative learning in real-world environments are also discussed.

## 1 Introduction and Background

In recent years, with the advancement of artificial intelligence (AI) and learning analytics technologies, an increasing number of researchers have sought to gain deeper insights into collaborative learning. Specifically, many researchers have explored the use of video data analysis to enhance their understanding of collaborative learning [1–3]. Gaze behaviour serves as a pivotal non-verbal cue in collaborative learning, offering insights into students' attention distribution, engagement levels, and group dynamics [4]. A growing number of research demonstrates that gaze-based metrics are strongly associated with collaboration quality and learning outcomes. Measures such as joint visual attention (JVA), gaze overlap, and gaze synchronisation have been linked to effective collaboration and higher learning gains [3, 5]. JVA, which reflects moments of shared focus on the same object or person, supports grounding, explanation, and joint problem-solving. Empirical studies in face-to-face and technology-enhanced settings consistently show that groups with higher levels of JVA perform better and demonstrate more productive interaction patterns [6, 7]. Beyond aggregate measures, where students direct their gaze also carries educational meaning. Peer-directed gaze is commonly associated with socially engaged, interactive collaboration, whereas prolonged gaze toward personal devices often reflects more individualised or task-oriented behaviour [5]. In one of the earliest classroom-based studies, Zhou et al. [5] identified four distinct gaze targets, including peers, laptops, tutors, and other objects, and showed that different gaze distributions corresponded to varying levels of shared understanding within groups. These findings highlight gaze as a behavioural marker that distinguishes collaborative knowledge construction from parallel individual work. Overall, gaze behaviours provide a direct window into attention, engagement, and coordination in collaborative learning. By capturing how learners visually align with peers and shared resources, gaze-based measures offer robust, theory-grounded indicators of collaboration quality that complement verbal and outcome-based analytics.

Despite its proven importance, capturing and interpreting gaze in authentic educational settings remains challenging. Traditionally, obtaining reliable gaze data has heavily relied on two primary methods, each with significant drawbacks. The first is manual video coding, where researchers review recordings frame-by-frame to label gaze targets, a process that is labour-intensive, time-consuming, and impractical for scaling or providing real-time feedback [3, 5, 8]. The second method involves wearable eye-tracking hardware, which, while precise, is often intrusive, costly, requires careful calibration, and is difficult to deploy at scale in dynamic classroom environments [9, 10].

To overcome these barriers, researchers have begun exploring vision-based methods to infer gaze from ordinary video footage [8]. Supervised machine learning (ML) approaches, such as convolutional neural networks (CNNs) or random forests, have shown promise but introduce new challenges [11]. These models need large amounts of labelled training data to learn how to predict where a person is looking. Obtaining such annotated datasets from classrooms is difficult [3]. Additionally, many machine learning models are trained on carefully curated datasets or controlled lab settings with limited diversity in lighting, occlusions, or group configurations, which reduces their robustness and cross-configuration robustness in real classroom environments [12, 13]. More advanced, Zhou and his colleagues [3] firstly proposed an automated framework. They proved the feasibility of utilising the object detection model to detect students and laptops, and used the spatiotemporal model [14] to predict the gaze target, achieving a moderate F1-score of 0.675 in detecting 2 types (students and laptops) of gaze behaviours in small-group collaborative learning settings. However, their testing is conducted using limited datasets (1000 frames or 6079 decisions) and simple students' positional configurations with no occlusion in the face during activities. Therefore, further evaluations utilising larger datasets with more complex scenarios are required to demonstrate the cross-configuration robustness of this approach [8]. In summary, prior work demonstrates steady progress toward scalable gaze detection in collaborative learning, yet challenges remain in reducing annotation requirements, handling context complexity and cross-configuration robustness. These limitations highlight the need for more adaptive, data-efficient and robust approaches while maintaining scalability and educational interpretability within multimodal learning analytics.

In response, we propose a scalable AI approach that integrates the YOLOE-26 model with text-prompt capability for education-related object detection [15], pre-trained YOLO11 for head detection and tracking [16], and the Gaze-LLE model (Gaze estimation via Large-scale Learned Encoders) for gaze target estimation [17]. The proposed approach outperforms traditional supervised machine learning baseline models (RF, CNN, MLP, SVM) and demonstrates consistent performance across varying collaborative group sizes and configurations. This work contributes to the field in the following key aspects: (1) By utilising foundation models, it eliminates the need for large manually labelled datasets for model training, making it more feasible for real-world implementation. (2) It bridges the cross-configuration robustness gap by demonstrating consistent performance across varying conditions, configurations and group sizes. (3) It provides empirical evidence of the approach's robustness, achieving higher detection accuracy compared to the prior study [3] to improve the state of the field. (4) The modular design of the pipeline enhances transparency at the system level, as each component performs a clearly defined function, allowing researchers to analyse and refine individual stages of the gaze inference process.

Towards building a gaze detector based on the gaps identified above, this work is organised around three research questions: (RQ1) whether an AI gaze detection pipeline can achieve high performance in automatically predicting students' gaze behaviours, (RQ2) how training data size affects the performance of supervised baselines relative to the proposed approach, and (RQ3) how robust the proposed approach is across different collaborative learning configurations.

## 2 Methodology

### 2.1 Models

To establish baseline performance for gaze behaviour detection (for comparison in RQ2 and RQ3), we implemented supervised machine learning models that require training on labelled gaze data. In our implementation, we utilised a YOLO11-based pose detector [18] to identify 17 body keypoints (e.g., eyes, nose, shoulders, etc.) for each person appearing in the video. Each detected person was then assigned a consistent person ID (e.g., 'Person 1', 'Person 2', ...) to maintain identity across frames. Using the annotated gaze behaviour dataset (see Section 2.2), we then trained four different classifiers to predict a student's gaze behaviour from the visual features. The classifiers included Random Forest (RF), Convolutional Neural Network (CNN), Multi-Layer Perceptron (MLP), and Support Vector Machine (SVM). These algorithms were used as they have demonstrated good performance in classification tasks in educational contexts [19] while no other competitive baseline exists. For each student in each video frame, the input to these models was the 17 body keypoint coordinates; the output was one of the gaze behaviour classes (gazing at a student, laptop, or other object). 5-fold cross-validation was applied to evaluate model accuracy.

Our proposed framework combines pretrained computer vision models with heuristic rules in a pipeline to detect gaze behaviours. The overall architecture of the system is depicted in Fig. 1. The system first detects and tracks students' heads using a YOLO11-based model, then estimates gaze points with the Gaze-LLE model and finally assigns each gaze to a target object or person through direct mapping. In parallel, a YOLOE-26 identifies potential targeted objects (like laptops, tablets and cell phones) in the scene.

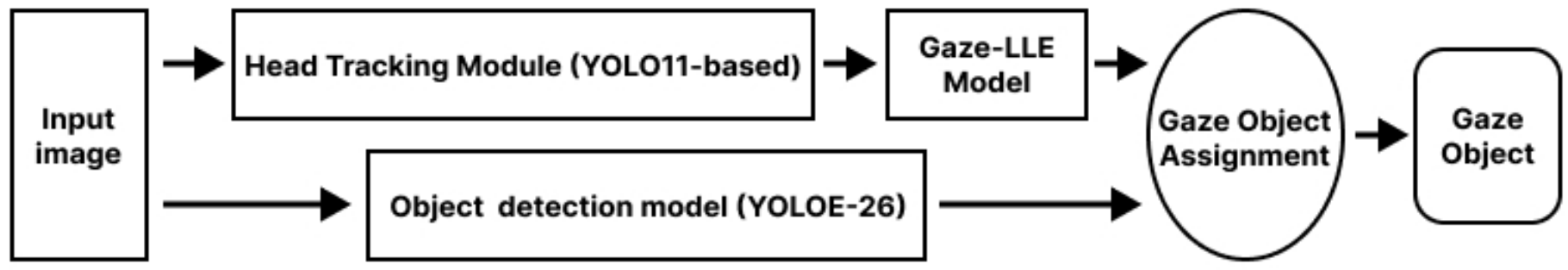

**Fig. 1.** Gaze detection pipeline.

**Head Tracking Module.** Head tracking is a fundamental step in detecting and analysing gaze behaviour as it provides bounding boxes for individuals appearing in video frames. These bounding boxes are important inputs to gaze estimation models. To detect head positions, the YOLO11m model was employed, which was trained using a custom dataset with 150 self-labelled images for whole head detection. The 150 images were sampled from pervious datasets (not the datasets used in this study) with similar educational contexts and data collection settings to ensure diversity in lighting conditions and student appearances. Annotation was performed using CVAT (cvat.org), with bounding boxes drawn around the full head region (crown to chin, ear to ear) for each visible student, following a written labelling protocol.

In the head tracking task for a specific group, a customised number (with 20 in this study) of random frames extracted from the video is used to train the head detection model for the purpose of students' ID tracking. Firstly, the number and position information of the head bounding boxes are derived from the pre-trained head detection model mentioned above. Then the 20 frames used for model training are filtered according to the preset number of group size and the distance between the table centre and the bounding boxes. The filtered head bounding boxes are then assigned with character labels (e.g., 'Person1', 'Person2', ...) in a clockwise direction based on the centre of the table. Person ID consistency is manually checked in this stage. Images with inconsistent IDs will be replaced and labelled automatically again until meeting the consistency requirement. Finally, the label information and head position information are generated into a training file for training the YOLO11-based head tracking model. The ratio of the training set to the validation set is 7:3. Finally, the head position information of the characters in all video frames is obtained from the head tracking model. The module framework is presented in Fig. 2.

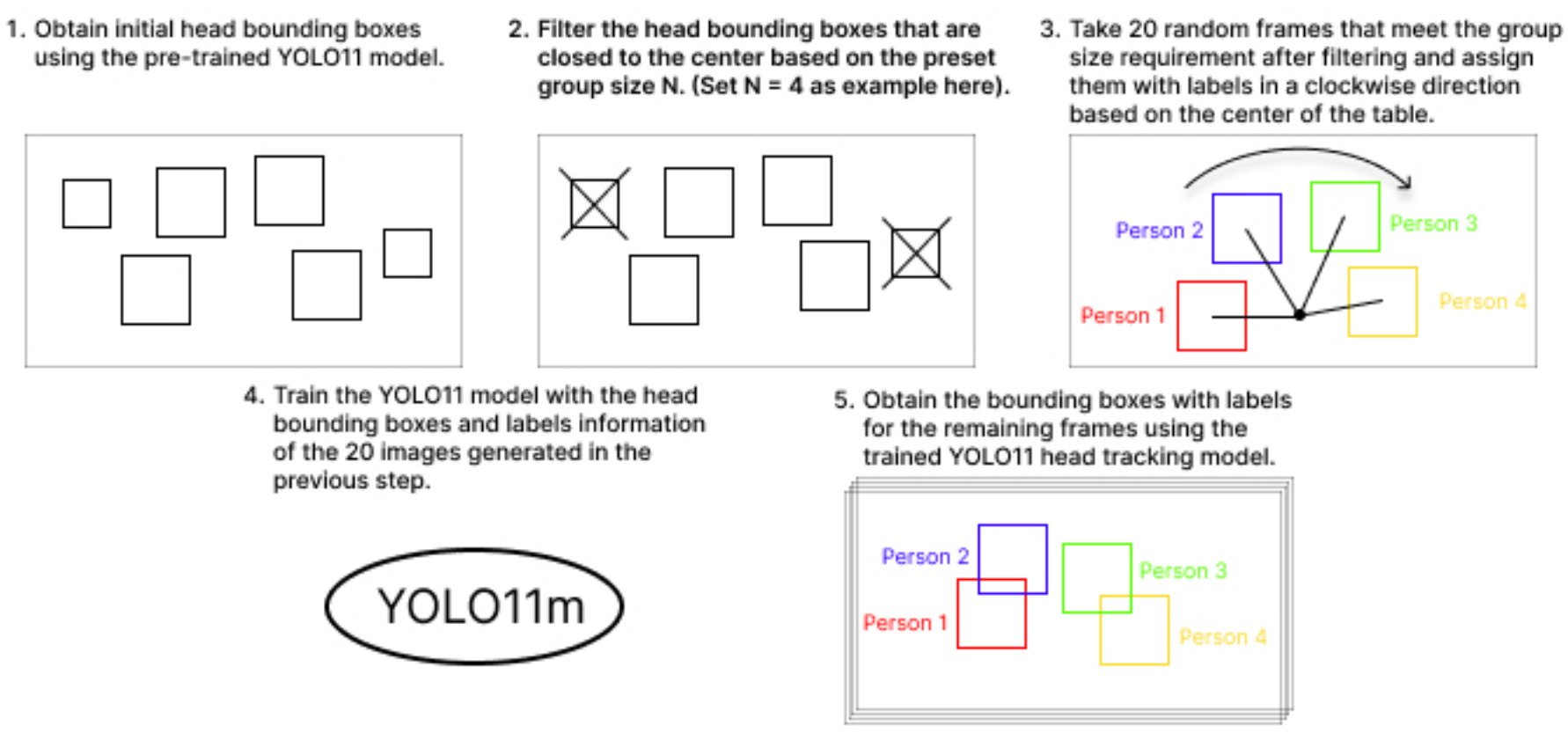

**Fig. 2.** The process of head tracking with an example group size of four.

**Gaze-LLE model for Gaze Estimation.** To estimate students' gaze targets from video frames, this study adopts Gaze-LLE, a foundation-model-based gaze target estimation approach that leverages large-scale pretrained visual representations [17]. Gaze-LLE formulates gaze estimation as a scene-level prediction task conditioned on the spatial location of an individual's head, enabling gaze inference without extensive labelled data. As shown in Fig. 3, the model uses a single frozen DINOv2 vision transformer to encode the entire scene, avoiding separate head, scene, or auxiliary encoders. A learned head-position prompt is injected into the corresponding spatial region of the scene feature map to condition the representation on the target individual. These head-conditioned features are then processed by a lightweight transformer-based decoder to generate a probabilistic gaze heatmap, from which the final gaze point is obtained by identifying the peak response. Within the proposed framework, Gaze-LLE serves as the core gaze estimation module, producing per-frame gaze predictions that are subsequently aligned with detected heads and objects to support socially and pedagogically meaningful analysis.

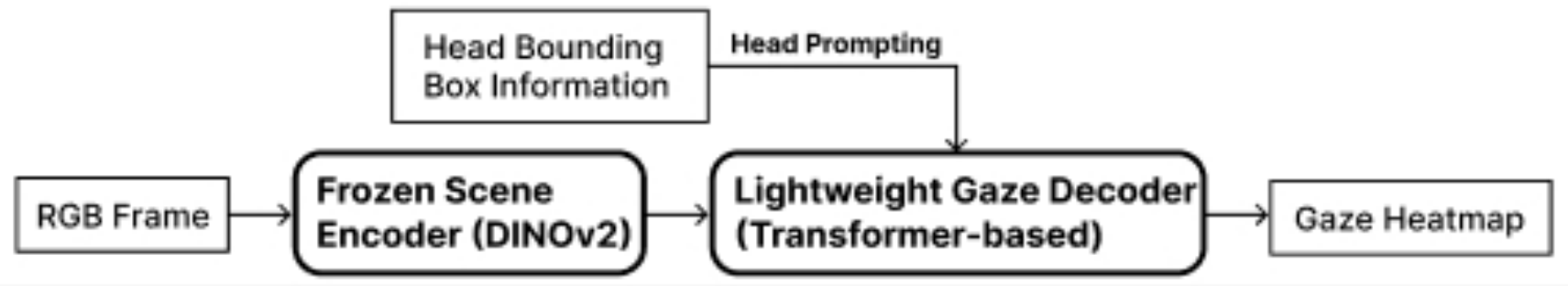

**Fig. 3.** Pipeline of the Gaze-LLE model.

**Object Detection and Gaze Point Assignment.** To identify task-relevant artefacts within collaborative learning scenes, the proposed framework employs YOLOE-26 for object detection [15]. YOLOE-26 is a pretrained, open-vocabulary object detection model built on the YOLO26 architecture, designed to support accurate detection without task-specific fine-tuning. In this study, the detector is configured to identify three commonly used digital learning artefacts: laptops, tablets, and phones. These objects represent typical focal points of student attention in face-to-face collaborative learning activities [5]. For each video frame, YOLOE-26 outputs bounding box information for detected objects. In parallel, the gaze estimation module produces predicted gaze point coordinates for each student. Gaze target assignment is performed by directly mapping the predicted gaze point to detected objects. Specifically, if the gaze point falls within the bounding box of a student's head/detected object, that student/object is assigned as the student's gaze target for the corresponding frame. In cases where a predicted point falls within multiple bounding boxes, the corresponding object whose centre point is closest to the predicted point will be assigned. When the gaze point does not intersect with any detected object's bounding box, no gaze target is assigned.

### 2.2 Dataset and Annotations

The dataset consists of video recordings of authentic face-to-face collaborative learning sessions collected over three academic years in a postgraduate course. Students worked weekly in groups of four or five around a T-shaped table to design technological solutions for educational challenges, primarily using a collaborative platform (Miro) on their laptops with instructors available for guidance. An Intel RealSense 435i camera was positioned at one end of the table to capture RGB video at 1080p and 30 FPS without intruding on the group's activity. Students can move freely during activities. Institutional ethical approval was obtained prior to data collection. All students provided informed written consent for video recording and were informed of their right to withdraw.

Six video clips were selected for analysis (two per year) in this study, including three clips with four-student groups (approximately 2.5 hours) and three with five-student groups (approximately 3 hours). To create a manageable annotation task, videos were sampled at one frame per second by extracting the first frame of each second. These frames were manually annotated for gaze behaviours.

Three gaze behaviour categories were defined following prior work: gazing at a student (S), gazing at a laptop (L), and gazing at other objects (O). Gazing at a student refers to instances where a student looks toward a peer within the same group, while gazing at a laptop captures attention directed at personal devices used for the collaborative task. The "other" category includes all remaining gaze behaviours (e.g., mobile phones or off-task directions). This categorisation has been shown to distinguish collaborative learning processes associated with different learning outcomes [3, 5].

Annotation was conducted by two researchers using the Computer Vision Annotation Tool (CVAT) (cvat.org). An initial set of 1,000 frames was double-coded to establish reliability, achieving high agreement (Cohen's Kappa = 0.784). The remaining frames were then annotated independently.

**Table 1.** The distribution of gaze behaviour annotations across different datasets.

| Dataset Name | Student (S) | Laptop (L) | Other (O) |
|---|---|---|---|
| Dataset 1 | 13926 | 3569 | 921 |
| Dataset 2 | 18522 | 7037 | 1471 |
| Dataset 3 (Dataset 1 + Dataset 2) | 32448 | 10606 | 2392 |

In total, 10,010 frames and 45,446 gaze behaviour decisions were annotated. Three datasets were constructed: Dataset 1 (five-student groups), Dataset 2 (four-student groups), and Dataset 3, which is a combination of Dataset 1 and Dataset 2, representing a mixed group size setting. (Table 1).

### 2.3 Evaluation Metrics

We evaluated model performance using standard classification metrics computed per gaze behaviour category and overall metrics as reported in similar prior studies [3, 14]. In particular, we report precision, recall and F1-score for each class (S, L, O), and an overall weighted average F1-score. A prediction is counted as correct if the model's assigned gaze category for a student matches the ground truth annotation for that student in that frame. We particularly focus on the F1 metric as an indicator of performance across classes as it accounts for both false positives and false negatives, whereas accuracy alone can be misleading in imbalanced datasets [20]. We also report statistical comparisons using the non-parametric Friedman test [21] to assess whether differences in performance across multiple models or contexts are significant.

### 2.4 Experiment Design

To answer the research questions, we designed three main experiments:
**Experiment 1 (Evaluating Proposed AI Approach – RQ1).** The proposed approach was applied to Dataset 3. Model performance was evaluated using class-wise and overall precision, recall, and F1-score, alongside a confusion matrix comparing predicted and ground-truth gaze behaviours. This experiment examines whether the proposed method can accurately detect gaze behaviours.
**Experiment 2 (Training Data Size vs. Performance – RQ2).** This experiment analyses how training data volume affects supervised baseline models and compares their performance with the proposed approach. Using Dataset 3, each supervised model (RF, CNN, MLP, SVM) was trained on increasing proportions of the dataset (10%–90%), with the remaining data used for testing. F1-score were recorded at each training level and compared against the constant performance of the proposed approach, which does not rely on task-specific training for gaze prediction purposes.
**Experiment 3 (Cross-Configuration Performance – RQ3).** To evaluate robustness across different group sizes and configurations, both the proposed approach and baseline models were tested in multiple collaboration configurations: groups of four (Dataset 1), groups of five (Dataset 2), and mixed group sizes (Dataset 3). Supervised models were trained and tested within each dataset using an 80/20 split. For the proposed approach, the model was directly applied to each dataset. Performance was compared using weighted F1-score, and the Friedman test was employed to assess statistical differences across configurations [21]. This experiment evaluates whether the proposed approach maintains stable performance as contextual complexity increases.

# 3 Results

## 3.1 Performance of the Proposed AI Approach (RQ1)

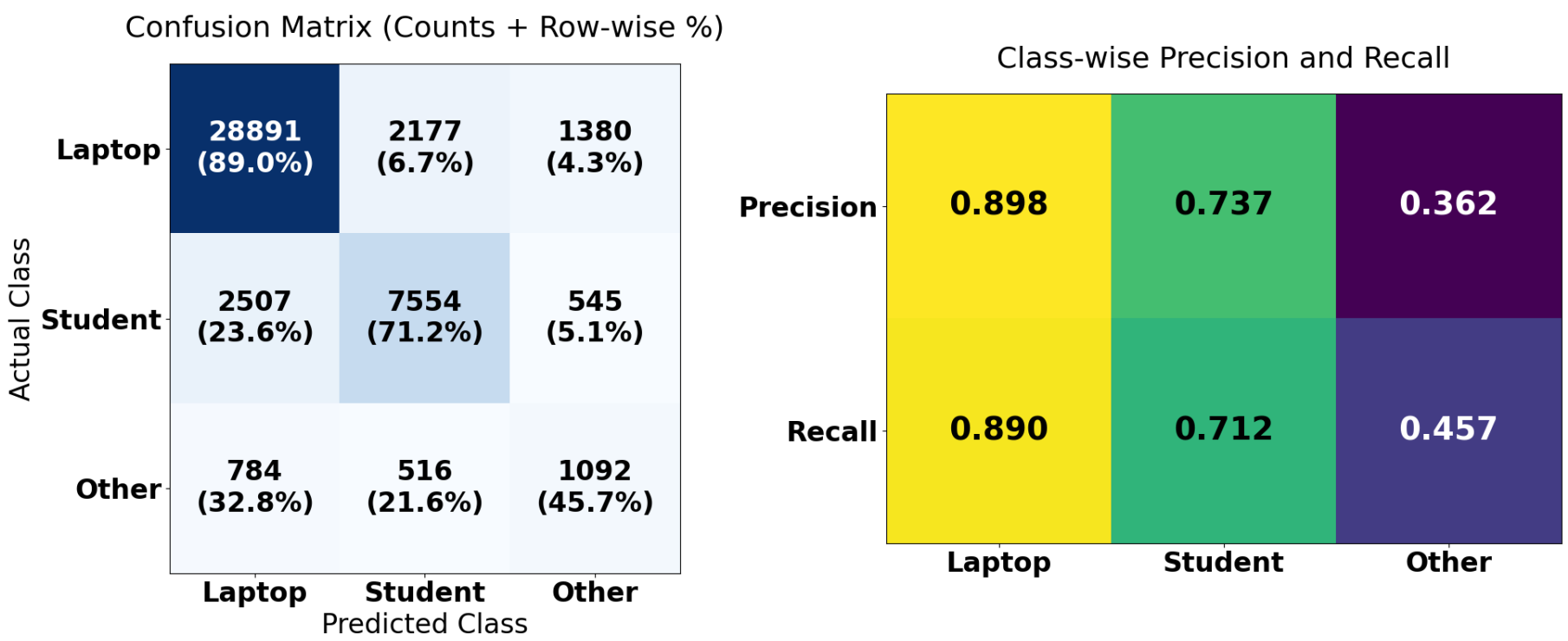

**Fig. 4.** Performance for the automated gaze detection model.

Fig. 4 shows the performance of the proposed approach in automatically detecting gaze behaviours in an authentic collaborative learning environment. The overall F1-score is 0.829. The precision for the laptop gaze behaviour class is 0.898, while the class recall is 0.890. Whereas the performance of detecting student-directed gaze behaviours was slightly lower at 0.737 and 0.712 in class precision and recall, respectively. It shows the effectiveness of the proposed approach to detect students' behaviours of gazing at laptops and gazing at students during the collaborative learning activities. However, the detection of gazing at other objects did not show promising performance, with only 0.362 and 0.457 in class precision and recall, respectively.

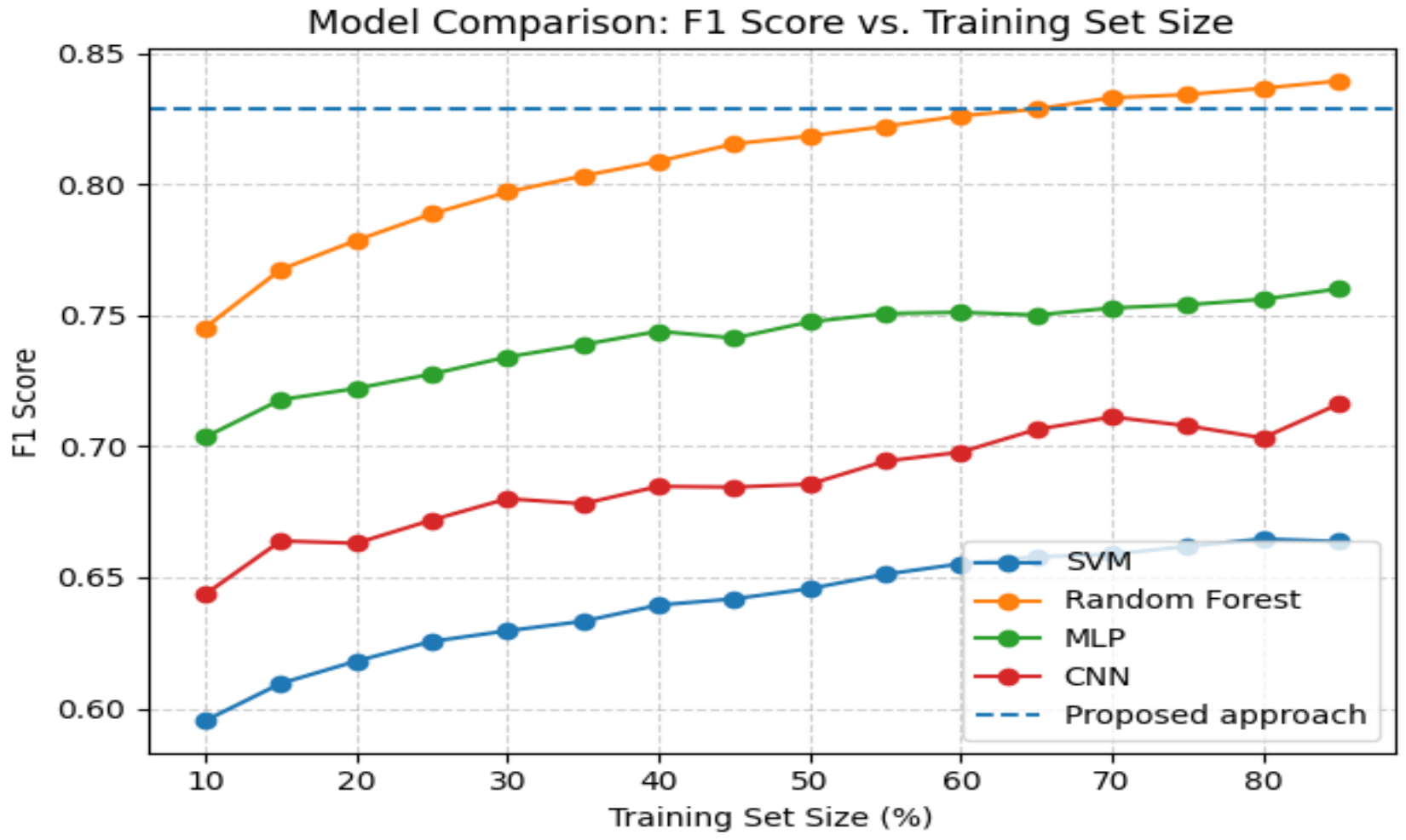

**Fig. 5.** Performance vs. training data size.

### 3.2 Effect of Training Data Size on Model Performance and Comparison (RQ2)

To investigate RQ2, we varied the amount of training data for the supervised baseline models and tracked their performance relative to the proposed approach. Fig. 5 plots the F1-score of each baseline model (RF, CNN, MLP, SVM) on the test set of Dataset 3 as the fraction of training data increases from 10% up to 90%. The horizontal red dashed line indicates the proposed model's F1 (0.829), which remains constant since it does not use training data for gaze prediction purposes. The results indicate that as the training dataset size increased, the performance of the supervised machine learning approaches improved accordingly. However, SVM, CNN, and MLP consistently underperformed compared to the proposed AI approach, regardless of training data size. Furthermore, the RF model outperformed the proposed approach when the training dataset reached about 66% of the data (about 29994 annotations).

### 3.3 Robustness Across Different Configurations (RQ3)

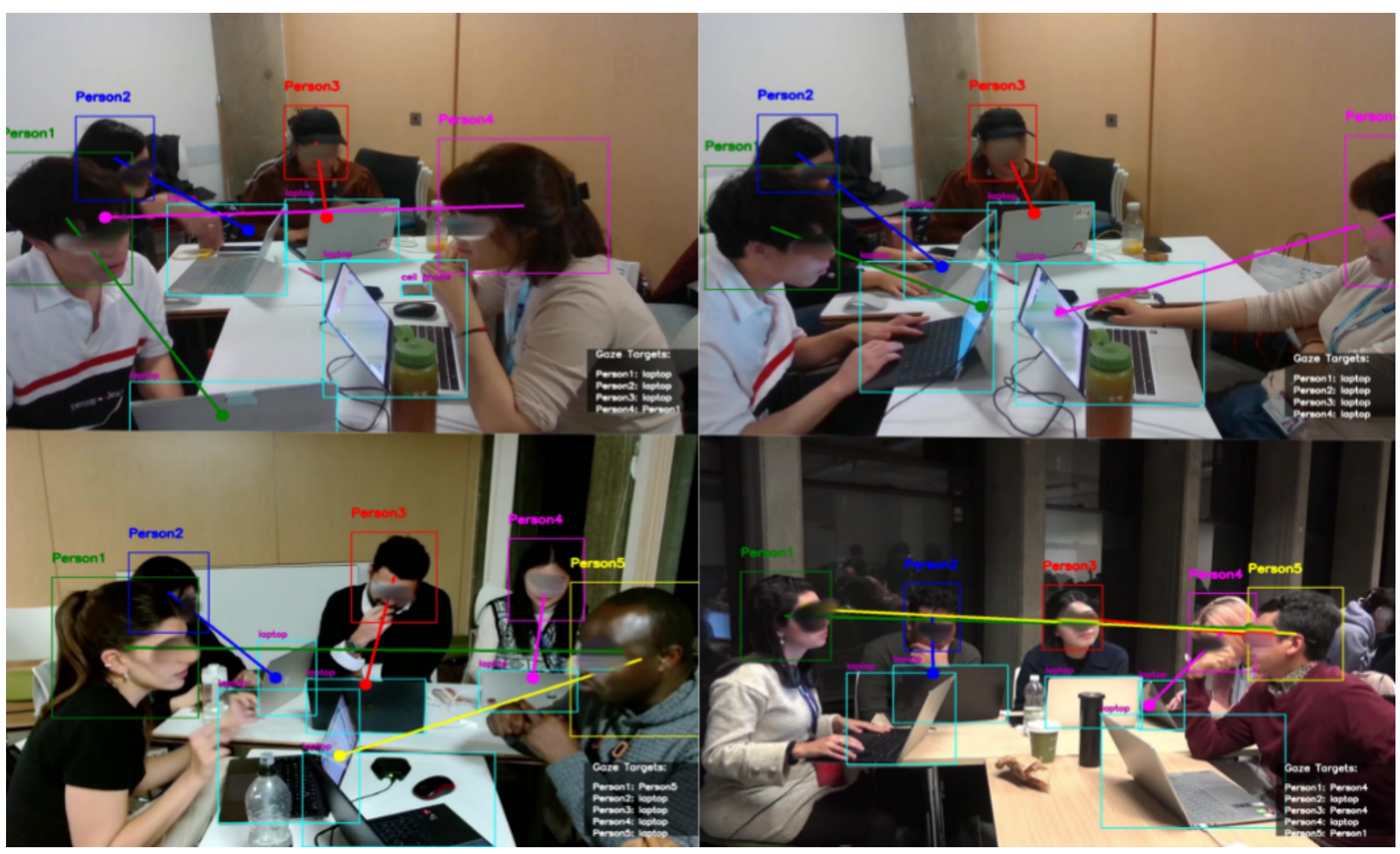

**Fig. 6.** Different configurations and conditions of collaborative learning context.

To address RQ3, we evaluated the proposed approach and baseline models across three collaboration configurations: groups of four (Dataset 1), groups of five (Dataset 2), and mixed group sizes (Dataset 3). Table 2 reports the weighted average F1-score for all models (proposed approach, RF, CNN, MLP, SVM). Supervised models were trained and tested within the same dataset using an 80/20 split to reflect deployment in a fixed context. Overall, supervised models performed best in the simplest setting (Dataset 1) and showed decreasing performance as context complexity increased. Statistical comparisons using the Friedman test confirmed significant differences across datasets [21]. While the proposed approach achieved significantly lower performance than supervised models in Dataset 1 ($p < .001$), it significantly outperformed them in Dataset 2 ($p < .001$) and achieved the second-highest F1-score in Dataset 3, trailing only the RF model ($p < .01$). Fig. 6 illustrates representative examples from the result, including different configurations, occluded faces, and posture changes. In contrast to supervised approaches, the proposed method maintained relatively stable performance across datasets, indicating consistent performance and robustness across varying configurations.

**Table 2.** The performance (F1-score) of different approaches across the three datasets.

| | RF | CNN | MLP | SVM | Proposed Approach |
|---|---|---|---|---|---|
| Dataset 1 (Group of four) | 0.877 | 0.860 | **0.886** | 0.822 | 0.856 |
| Dataset 2 (Group of five) | 0.759 | 0.645 | 0.701 | 0.591 | **0.810** |
| Dataset 3 (Mixed group size) | **0.837** | 0.703 | 0.756 | 0.665 | 0.829 |

## 4 Discussion

The results demonstrate that the proposed AI approach can reliably detect educationally meaningful gaze behaviours in real-world collaborative learning settings. Firstly, in relation to RQ1, the proposed approach achieved an overall F1-score of 0.829, demonstrating strong performance in detecting laptop-directed gaze (Precision = 0.898, Recall = 0.890) and moderate performance for student-directed gaze (Precision = 0.712, Recall = 0.737). In contrast, performance for the "Other" category was lower. A likely reason is that "Other" encompasses diverse and less distinct targets, which weakens the rule-based mapping process. Despite accuracy not providing a complete picture of the model's performance [20], a number of studies reported only accuracy rather than a more comprehensive indicator, such as the overall F1-score. For example, Zhou [3] presented an approach for detecting two types (peer and laptop) of students' gaze behaviours in a similar context, achieving an accuracy of 66.57% and an F1-score of 0.675. Furthermore, given the limited number of studies focused on detecting gaze behaviours in such a complex context, we also compared our approach with baseline models for similar tasks. Muller [22] attempted to detect participants' eye contact in group meetings using videos from eight cameras with a model that required no training, achieving an accuracy of 54%. Wu [23] proposed an unsupervised approach to detecting eye contact in real-world videos, achieving 71.88% accuracy. Stiefelhagen [24] developed a method for detecting the focus of attention in meetings by analysing video data from a panoramic camera, reaching an accuracy of 73.9%. Chong [14] inferred shared attention by detecting gaze behaviours in social scenes using video data, achieving 83.3% accuracy. It is worth noting that while Chong's [14] approach can infer the general area that was gazed at, it cannot determine which specific objects were gazed at.

Secondly, addressing RQ2, all supervised baseline models improved as training data increased, with Random Forest achieving the strongest performance among them. However, CNN, MLP, and SVM models consistently underperformed relative to the proposed approach, even when trained on most of the dataset. While RF surpassed the proposed method at higher data volumes, this required substantial manual annotation effort. These results highlight the data efficiency of the proposed approach, which achieves competitive performance without requiring extensive training data. This finding aligns with prior research showing that limited and context-specific datasets constrain the generalisability of supervised models in educational settings [25].

Thirdly, with respect to RQ3, cross-configuration robustness is considered a challenge in learning analytics [26]. Machine learning algorithms map predictor variables to outcomes by analysing data patterns in historical datasets and inferring underlying rule sets. However, due to the dynamic and non-deterministic nature of the learning process, historical datasets can rarely represent all possible situations in educational contexts [26]. Although performance in groups of five (F1 = 0.810) was slightly lower than in groups of four (F1 = 0.856), it remained ideal compared to previous work (F1 = 0.675) [3] and did not show a substantial decline. This stability across group sizes indicates that the proposed method is not highly sensitive to specific group configurations. In contrast, supervised machine learning models showed marked performance drops in more complex contexts, suggesting that even with sufficient training data, they struggle to generalise across settings. Overall, the proposed AI approach demonstrated more consistent performance across datasets, indicating superior cross-configuration robustness and supporting its use for researchers to analyse gaze behaviours in diverse collaborative learning environments.

### 4.1 Educational Practice

Beyond the technical performance, the proposed approach offers significant practical advantages for its application in real-world educational settings. These advantages can be broadly categorised into scalability, data efficiency, and the potential for near-real-time learning support, which collectively address long-standing barriers to the adoption of advanced analytics in collaborative learning.

A primary barrier to the widespread adoption of sensor-based learning analytics is the high cost and complexity of the required equipment [27]. Previous studies have relied on specialised sensors such as intrusive eye trackers [10] to capture gaze behaviours during learning. While precise, these tools impose substantial financial burdens and require complex calibration and installation, making them impractical for large-scale, sustained use in authentic classrooms. In contrast, our approach requires only a single standard 2D RGB camera, a resource that is increasingly ubiquitous and affordable in educational institutions. This shift from specialised hardware to commodity hardware dramatically lowers the entry cost, enabling the scalable deployment of gaze behaviour analysis across multiple classrooms and institutions simultaneously.

The data efficiency of our proposed approach is a key enabler for practical applications. As demonstrated in RQ2, while supervised models like Random Forest can achieve high accuracy, they require a substantial volume of manually annotated data. In this case, approximately 29539 annotations (65% of the dataset) are required to surpass the performance of our annotation-free method. The reliance on extensive, context-specific training data is a major bottleneck, as the process of human annotation is time-consuming, expensive, and impractical for providing real-time feedback [5, 28]. Our approach bypasses this requirement by leveraging pre-trained foundation models.

This makes it feasible to deploy the system in new collaborative learning environments, facilitating immediate and adaptable applications.

Furthermore, the automation and efficiency of our method unlock the potential for near-real-time analysis of collaborative learning processes. Prior research has established a clear connection between gaze behaviours and collaboration quality and learning outcomes accordingly [4, 5, 10]. However, the absence of automated detection has meant that analyses were retrospective, conducted long after the learning activity concluded. This delay severely limits the utility of the insights for timely intervention. Our system, which does not rely on post-hoc annotation, has the potential to be integrated into a pipeline capable of providing educators and learners with immediate feedback. We envision two potential feedback modalities. First, a session-level summary generated within minutes of activity end, displaying the proportions of different gaze behaviour per student for the instructor. Second, similar to [28] but real-time, an ambient indicator or panel signals when a group's collective peer-directed gaze drops sharply, prompting a brief facilitation move for student reflection and regulation. This capacity for timely support addresses a critical gap in learning analytics by moving from post-hoc analysis to near-real-time scaffolding, potentially enhancing the effectiveness of collaborative learning interventions [29].

### 4.2 Limitations and Future Work

Despite the promising results, this study has several limitations that should be noted. A limitation of the baseline comparison is the asymmetry in input features. Supervised models receive 17 body keypoint coordinates as input, whereas the proposed pipeline processes the full RGB scene through deep visual encoders. This disparity means the performance gap should not be interpreted as a fundamental ceiling of supervised machine learning. Rather, it highlights a data-annotation cost, which means extracting richer visual features for supervised training would itself require substantial labelling effort, reinforcing the data-efficiency argument. Future work could explore whether richer supervised inputs (e.g., cropped head images) narrow the gap. Secondly, the weak performance in the other category warrants closer examination. This class accounts for only about 5% of all annotations, creating a class imbalance. Educationally, 'Other' captures gaze directed at tutors, shared whiteboards, or off-task directions, which can signal disengagement or active tutor-seeking. Future work should pursue increasing annotated 'Other' instances and splitting the category into subclasses such as tutor-directed vs. off-task. Thirdly, the current approach considers only three types of gaze behaviours, gazing at students, laptops, and other objects. While these categories can capture important aspects of collaboration, such as resource management and peer interactions, they may not fully represent the complexity of gaze dynamics in face-to-face learning environments. Expanding the classification to incorporate additional gaze behaviours, such as gazing at tutors or shared materials, could provide deeper insights into collaborative learning processes. Finally, a promising future direction is to integrate multimodal large language models (LLMs) into the gaze target assignment step, replacing direct geometric mapping with a more context-aware, adaptive system. LLM-based models can draw on multimodal cues such as facial expressions, postures, and task information to infer targets more accurately [29], allowing richer reasoning over gaze behaviour and better handling of multiple targets or subtle social dynamics. Such a shift would further improve generalisability across various settings.

# 5 Conclusion

In this paper, we proposed a scalable AI approach to automatically detect students' gaze behaviours in real-world collaborative learning activities by integrating pretrained foundation models. Given the approach's demonstrated accuracy and cross-configuration robustness across different collaborative learning settings, we conclude that it holds strong potential for application in diverse collaborative learning environments. This method offers a scalable solution for analysing non-verbal behaviours in the collaborative learning process. Furthermore, the findings have important implications for enhancing real-time automated feedback interventions to support and improve collaborative learning experiences.

**Acknowledgments.** The second author's research is co-funded by the European Commission's project "Teacher-AI Complementarity (TaiCo)" (Project ID: 101177268).